\documentclass[letterpaper,10pt,conference]{ieeeconf}

\usepackage{amsmath,amssymb,mathtools}
\usepackage{algorithm}
\usepackage{algorithmic}
\usepackage{graphicx}
\usepackage{booktabs}
\usepackage{cite}
\usepackage{url}

\newtheorem{proposition}{Proposition}
\newtheorem{theorem}{Theorem}

\newcommand{\R}{\mathbb{R}}

\newcommand{\KL}{\mathrm{KL}}
\newcommand{\MMD}{\mathrm{MMD}}

\IEEEoverridecommandlockouts
\begin{document}

\title{Rapid Mean-Field Formation Transport with a Shared-Representation Neural Fokker--Planck Controller}

\author{Zejian Zhou$^{1}$ and Hao Xu$^{2}$%
\thanks{$^{1}$Department of Electrical Engineering and Computer Science, University of Wyoming, Laramie, WY 82071, USA.}%
\thanks{$^{2}$Department of Electrical and Biomedical Engineering, University of Nevada, Reno, Reno, NV 89557, USA.}%
\thanks{Accepted for publication in the 2026 IEEE Conference on Decision and Control (CDC).}%
\thanks{\copyright\ 2026 IEEE. Personal use of this material is permitted. Permission from IEEE must be obtained for all other uses, in any current or future media, including reprinting/republishing this material for advertising or promotional purposes, creating new collective works, for resale or redistribution to servers or lists, or reuse of any copyrighted component of this work in other works.}%
}

\maketitle

\begin{abstract}
Mean-field control describes the motion of a large agent population through its density, but solving the associated planning equations can be costly for online control. We propose a shared-representation neural Fokker--Planck controller for transporting a swarm toward prescribed formations. A shared neural network represents the population density and feedback control law through two output heads. Offline training combines Fokker--Planck and initial-condition residuals with formation costs evaluated on simulated particle trajectories and a control-effort penalty. Each agent then evaluates the trained feedback law using its own state and time. For a periodic state domain, we establish well-posedness of the induced density evolution and bound the density approximation error by the PDE and initial-condition residuals. One-dimensional transport and two-dimensional formation experiments illustrate the trade-off between terminal accuracy, control effort, and computation time. The formation experiments show lower transport errors than the potential-field and consensus baselines, while an architecture comparison supports the use of shared features for density and control.
\end{abstract}


\section{Introduction}
Large-scale robotic formation control is becoming increasingly important for tasks requiring scalable and robust coordination \cite{DorigoTheraulazTrianni2021}. Classical approaches include potential-field control \cite{RimonKoditschek1992} and consensus-based coordination \cite{RenBeard2008}, but distributional objectives under stochastic disturbances motivate a population-level description. Mean-field formulations are attractive in this regime because they describe the swarm through a population density and interpret formation synthesis as probability transport \cite{LasryLions2007,Bensoussan2013}. This formulation characterizes the spatial distribution of large agent populations under stochastic disturbances and expresses formation objectives directly in terms of the population density.

Numerical methods for coupled mean-field partial differential equations (PDEs) include policy iteration with repeated forward--backward solves \cite{CacaceCamilliGoffi2021} and alternating optimization of population and control approximations \cite{lin2021alternating}. When these computations are repeated online, their cost can limit feedback implementation under restricted computational resources. Optimization based on particle simulations evaluates formation objectives directly, but does not explicitly impose the Fokker--Planck equation on an auxiliary density approximation. The control synthesis problem is therefore to incorporate the probability-transport dynamics and formation objectives into a feedback law that can be evaluated with low online computational cost.

This paper jointly parameterizes a decentralized feedback law and an auxiliary density approximation using a neural network, defined as a finite composition of affine maps and componentwise nonlinear functions. A common transformation of state and time supplies intermediate coordinates to separate density and control output maps. The coefficients are computed by minimizing an objective that combines density residuals, formation mismatch, and control effort. This numerical parameter optimization, termed training or learning, is performed offline, before online feedback execution (deployment). Formation mismatch is evaluated on simulated closed-loop particle trajectories (rollouts); their use in parameter optimization is termed rollout supervision. During online operation, the coefficients remain fixed, and each agent evaluates the feedback law using only its current state and time.

The contributions are summarized as follows.
\begin{itemize}
\item A common state-time parameterization is constructed for the feedback law and density approximation. The shared coefficients couple the two maps during offline optimization, while online control requires only evaluation of the feedback map at each agent's state and the current time.
\item A joint synthesis objective combines Fokker--Planck and initial-condition residuals with particle-based formation costs and control effort. The residual terms promote consistency with the density dynamics, while the particle-based terms penalize deviations from prescribed formations at selected times.
\item For the periodic-domain model, well-posedness of the induced density evolution and a residual-to-density error bound are established under the stated assumptions. The bound quantifies consistency between the density approximation and the exact density induced by the synthesized feedback law.

\end{itemize}

\section{Background}
\subsection{Mean-Field Control and Probability Transport}
Mean-field formulations describe the collective behavior of large agent populations through a state distribution \cite{LasryLions2007}. For stochastic agent dynamics, the evolution of this distribution is characterized by a controlled Fokker--Planck equation \cite{Bensoussan2013}. Formation control can therefore be formulated as steering an initial population density toward a prescribed target distribution subject to drift and diffusion. Decentralized control under population coupling has been studied for large stochastic multi-agent systems \cite{LiZhang2008}, with applications to swarm tracking developed within mean-field game formulations \cite{zhou2021large}. Mean-field games seek consistency between individually optimal control laws and the resulting population distribution. The formulation considered here instead minimizes a common population objective, while retaining the controlled Fokker--Planck description of density evolution.

Numerical approaches to mean-field planning and mean-field games include policy iteration \cite{CacaceCamilliGoffi2021}, proximal-gradient optimization \cite{YuLaiLiOsher2024MFP}, and alternating updates of neural density and control approximations \cite{lin2021alternating}. Coupled approximations of the control law, value function, and population density have also been used to construct mean-field-game controllers for large multi-agent systems \cite{ZhouXu2022TCYB}. For methods that solve the coupled optimality equations separately for each planning instance, changes in the task specification require additional computation. This requirement can limit their use when control inputs must be computed within a restricted online computational budget.

\subsection{Physics-Informed Solvers for Mean-Field Games}
Physics-informed approximation incorporates governing-equation residuals into parameter optimization. Applications to mean-field games include spatiotemporal physics-informed neural operators \cite{LiuChenDi2024PINO} and unsupervised solution-operator learning \cite{HuangLai2025USOLMFG}. These approaches approximate solution fields or mappings from problem data to solutions while incorporating the governing equations into the approximation procedure.

The coupling between density evolution and individual optimality remains part of the mean-field game formulation. Physics-informed neural network (PINN) methods address this coupling through residual-based optimization of network coefficients. Operator approximations can reduce the computation required for new problem instances by performing parameter optimization offline; their computational requirements depend on the class of problems represented. This paper considers a prescribed population-level formation objective and constructs an explicit feedback law through joint offline optimization of density and control approximations. During online operation, each agent evaluates this law at its current state and time with fixed coefficients, without solving a coupled forward--backward system.

\section{Problem Formulation}
\subsection{Formation-Control Problem}

Let $\Omega$ denote the state domain, $T>0$ the time horizon, and $x_i(t)\in\Omega$ the state of agent $i$, for $i=1,\dots,N$. The continuum model and analysis use the periodic domain $\Omega=\mathbb T^d$. Bounded Euclidean boxes, also denoted by $\Omega$ when discussing numerical computations, are treated separately through the boundary rules specified in Section~\ref{sec-controller}. The swarm distribution is represented by the empirical measure
\[
\mu_t^N=\frac{1}{N}\sum_{i=1}^N \delta_{x_i(t)},
\]
where $\delta_x$ denotes the Dirac measure concentrated at $x$, so $\mu_t^N$ records the swarm shape at time $t$. Let $C^{\mathrm{tar}}=\{c_j\}_{j=1}^M\subset\Omega$ denote a set of target points. The target measure
\[
\nu^{\mathrm{tar}}=\frac{1}{M}\sum_{j=1}^M \delta_{c_j}
\]
specifies the desired allocation of mass among these points. More generally, target samples can describe a continuous spatial distribution around the desired geometry. The control objective is to drive $\mu_t^N$ toward the prescribed target distribution at selected times while limiting control effort.

On the periodic domain, each agent obeys the stochastic dynamics
\begin{equation}
\begin{aligned}
dx_i(t)
&=
\bigl(f(x_i(t))+g(x_i(t))u(x_i(t),t)\bigr)\,dt
\,+\,\sigma\,dW_i(t),
\end{aligned}
\label{eq-sde}
\end{equation}
where $f:\Omega\to\R^d$ and $g:\Omega\to\R^{d\times m}$ are known, $u(x,t)\in\R^m$ is the common feedback law, and $\sigma>0$ is the diffusion coefficient. Equation~\eqref{eq-sde} is interpreted periodically. The $W_i$ are independent standard $d$-dimensional Wiener processes, independent of the initial states, which are independently sampled from $\rho_0$. The standing analytical assumptions are $f\in W^{1,\infty}(\Omega;\R^d)$, $g\in W^{1,\infty}(\Omega;\R^{d\times m})$, and $\rho_0\in L^2(\Omega)$ with $\rho_0\ge0$ and $\int_\Omega\rho_0\,dx=1$. The spaces $H^1$ and $W^{1,\infty}$ are the usual periodic Sobolev spaces, and $H^{-1}=(H^1)^*$.

Fix a control bound $M_u>0$. The admissible class $\mathcal U_{\rm ad}$ consists of deterministic feedback laws that are measurable in time, Lipschitz continuous in state uniformly in time, and satisfy $\|u\|_{L^\infty(\Omega\times(0,T))}\le M_u$. Each such law yields well-posed agent dynamics. Population coupling enters through the common design objective rather than an explicit density-dependent drift.

\subsection{Mean-Field Control Formulation}
For a fixed $u\in\mathcal U_{\rm ad}$, independent of the population size $N$, let $\rho^u(x,t)$ denote the probability density of the common state law. Write $\rho=\rho^u$ when the feedback is understood. The empirical measure $\mu_t^N$ is a random finite-sample representation of this law. Independence gives, for each fixed $t$ and continuous test function $\varphi$,
\[
\int_\Omega\varphi\,d\mu_t^N
=\frac{1}{N}\sum_{i=1}^N\varphi(x_i(t))
\xrightarrow[N\to\infty]{\mathrm{a.s.}}
\int_\Omega\varphi(x)\rho(x,t)\,dx.
\]
Thus $\rho(\cdot,t)\,dx$ is the weak limit of $\mu_t^N$. Densities at different evaluation times are time slices of this same evolving law, corresponding to the empirical measures at those times.

For a smooth periodic test function $\varphi$, define the controlled diffusion generator
\[
\mathcal L_u\varphi(x,t)
=
\nabla\varphi(x)\!\cdot\!(f(x)+g(x)u(x,t))
\,+\,
\frac{\sigma^2}{2}\Delta\varphi(x).
\]
Then It\^o's formula applied to \eqref{eq-sde} and averaged over the swarm gives
\[
\begin{aligned}
d\!\int_\Omega \varphi(x)\,\mu_t^N(dx)
&= 
\int_\Omega \mathcal L_u\varphi(x,t)\,\mu_t^N(dx)\,dt \\
&\quad
+\,
\frac{\sigma}{N}\sum_{i=1}^N \nabla\varphi(x_i(t))\cdot dW_i(t).
\end{aligned}
\]
By It\^o's isometry, the stochastic integral over $[0,t]$ has second moment at most $\sigma^2T\|\nabla\varphi\|_\infty^2/N$ for $t\le T$. Taking expectations gives
\[
\frac{d}{dt}\int_\Omega \varphi(x)\rho(x,t)\,dx
=
\int_\Omega \mathcal L_u\varphi(x,t)\rho(x,t)\,dx,
\]
for all smooth periodic test functions $\varphi$ and almost every $t$, which is the weak form of the controlled Fokker--Planck equation. Hence formation control becomes a probability-transport problem for the swarm density, and the shared feedback law in \eqref{eq-sde} induces
\begin{equation}
\partial_t \rho + \nabla\!\cdot\!\bigl((f+gu)\rho\bigr)-\frac{\sigma^2}{2}\Delta \rho = 0
\quad \text{in } Q=\Omega\times(0,T),
\label{eq-fpe}
\end{equation}
with initial density $\rho(\cdot,0)=\rho_0$. The periodic model requires no boundary flux condition. Numerical computations on bounded boxes use an auxiliary flux penalty and particle projection, which are not covered by the periodic-domain analysis. Let $\mathcal T_s=\{t_k\}_{k=1}^{N_s}$, with $0<t_1<\cdots<t_{N_s}\le T$, denote the formation-evaluation times, and let $\{\rho_k^{\mathrm{tar}}\}_{k=1}^{N_s}$ denote the prescribed target densities at those times. A finite target measure $\nu^{\mathrm{tar}}$ is atomic and has no Lebesgue density; a continuous target can instead be specified by smooth profiles around the target points. Empirical measures of independent, identically distributed samples from $\rho_k^{\mathrm{tar}}$ converge weakly almost surely to $\rho_k^{\mathrm{tar}}(x)\,dx$ as their sample size increases. For static formations, one may take $\rho_k^{\mathrm{tar}}\equiv \rho^{\mathrm{tar}}$ for all $k$, whereas for staged transport tasks the target density may vary with $k$. The mean-field formation-transport problem is therefore to choose a shared feedback law $u$ such that the induced density $\rho(\cdot,t_k)$ approaches $\rho_k^{\mathrm{tar}}$ at the selected times while maintaining moderate control effort. The target geometry is imposed by penalizing the mismatch between simulated particle distributions and target formations at the times in $\mathcal T_s$.

\section{Shared-Representation Neural Fokker--Planck Controller}
\label{sec-controller}

\subsection{Shared-Representation Density-Control Parameterization}
The approximations $\rho_\theta$ and $u_\theta$, parameterized by the coefficient vector $\theta$, are coupled through the transport term
\[
\bigl(f(x)+g(x)u_\theta(x,t)\bigr)\rho_\theta(x,t).
\]
The shared feature map $T_\theta \colon \Omega\times[0,T]\to\R^p$ computes $p\in\mathbb N$ intermediate coordinates used by both outputs. The density head is the scalar output map $S_\theta:\R^p\to\R$, and the control head is the vector output map $U_\theta:\R^p\to\R^m$. They use the common coordinates $h_\theta$ to compute the raw scalar $\tilde\rho_\theta(x,t)$ and feedback input $u_\theta(x,t)$, respectively. With $\theta$ collecting all adjustable coefficients,
\begin{equation}
h_\theta(x,t)=T_\theta(x,t), \qquad
\tilde \rho_\theta=S_\theta(h_\theta), \qquad
u_\theta=U_\theta(h_\theta).
\label{eq-shared_param}
\end{equation}
The raw output $\tilde\rho_\theta$ may have either sign and is not itself a probability density. The common map followed by the two output maps defines the two-head architecture, termed the shared-representation neural Fokker--Planck controller (SR-NFPC).
The density output is normalized at each time by
\begin{equation}
\rho_\theta(x,t)=\frac{\operatorname{softplus}(\tilde\rho_\theta(x,t))+\varepsilon}
{\int_\Omega \left(\operatorname{softplus}(\tilde\rho_\theta(\xi,t))+\varepsilon\right)d\xi},
\label{eq-density_norm}
\end{equation}
where $\operatorname{softplus}(z)=\log(1+\exp(z))$ and $\varepsilon>0$ is a fixed numerical offset. Since $|\Omega|<\infty$, exact evaluation of the denominator gives $\rho_\theta>0$ and $\int_\Omega\rho_\theta(\cdot,t)\,dx=1$; numerical quadrature introduces an integration error. The term learned density refers to the approximation $\rho_\theta$ obtained by parameter optimization. Equations \eqref{eq-shared_param} and \eqref{eq-density_norm} define a coupled map
\[
(x,t)\mapsto \bigl(\rho_\theta(x,t),u_\theta(x,t)\bigr)
\]
with shared intermediate coordinates. The control output satisfies $\|u_\theta\|_{L^\infty(Q)}\le M_u$. Smooth nonlinearities provide the derivatives required for collocation and yield a feedback law that is Lipschitz continuous in state on the compact state-time domain. For the periodic analysis, both approximations are required to be periodic in state, so $u_\theta\in\mathcal U_{\rm ad}$. Online execution requires only the composition $u_\theta=U_\theta\circ T_\theta$.

\subsection{Physics-Informed Density Residuals}
Density residuals measure defects in the controlled Fokker--Planck equation and its initial condition. Define the interior operator
\[
\mathcal F[\rho_\theta,u_\theta]
=
\partial_t\rho_\theta
\,+\,\nabla\!\cdot\!\bigl((f+g u_\theta)\rho_\theta\bigr)
\,-\,\frac{\sigma^2}{2}\Delta\rho_\theta,
\]
so the formal mean-field model \eqref{eq-fpe}, together with $\rho(\cdot,0)=\rho_0$, requires
\[
\begin{aligned}
\mathcal F[\rho,u]&=0
&& \text{in } Q, \\
\rho(\cdot,0)-\rho_0&=0.
\end{aligned}
\]
The dynamics and initial-condition residuals entering the formal analysis are
\begin{align}
\varepsilon_{\mathrm{dyn}}^2
&= 
\int_0^T
\left\|
\mathcal F[\rho_\theta,u_\theta](\cdot,t)
\right\|_{H^{-1}(\Omega)}^2\,dt,
\label{eq-eps_dyn}
\\
\varepsilon_0^2
&=
\left\|
\rho_\theta(\cdot,0)-\rho_0
\right\|_{L^2(\Omega)}^2.
\label{eq-eps_initial}
\end{align}
For a square-integrable residual, the continuous norm satisfies
\[
\varepsilon_{\mathrm{dyn}}
\le C_\Omega\|\mathcal F[\rho_\theta,u_\theta]\|_{L^2(Q)},
\]
where $C_\Omega$ is the norm of the embedding $L^2(\Omega)\hookrightarrow H^{-1}(\Omega)$. Collocation approximates the squared $L^2(Q)$ norm by residual evaluations at sampled points. This finite-sample quantity does not itself bound $\varepsilon_{\mathrm{dyn}}$ without an additional sampling-error estimate. All derivatives act on the normalized density in \eqref{eq-density_norm}, including the time-dependent denominator.

For the bounded box $\Omega$ used in numerical computations, let $\boldsymbol n$ denote the outward unit normal on $\partial\Omega$. The auxiliary boundary penalty is
\begin{equation}
\varepsilon_{\mathrm{flux}}^2
=
\left\|
\left(
(f+g u_\theta)\rho_\theta
-\frac{\sigma^2}{2}\nabla\rho_\theta
\right)\!\cdot\boldsymbol n
\right\|_{L^2(\partial\Omega\times(0,T))}^2,
\end{equation}
which penalizes departures from zero normal probability flux. For the periodic model, set $\varepsilon_{\mathrm{flux}}=0$. The bounded-domain penalty does not impose an exact boundary condition.

Minimizing these residuals promotes consistency with the transport-diffusion equation and initial distribution. These constraints alone do not select a target formation, since different feedback laws can induce different density evolutions satisfying the same PDE model and initial condition.

\subsection{Formation Matching by Particle Simulation}
Formation costs evaluated on simulated particle distributions supply geometric information absent from the PDE residuals. These simulations use the same feedback law as online operation.

For the bounded-box numerical model, use $n$ simulated particles initialized independently from $\rho_0$. Let $L$ be the number of integration steps, $\Delta t=T/L$, and $s_\ell=\ell\Delta t$. The projected Euler--Maruyama update is
\begin{equation}
x_i^{\ell+1}
\!=\!
\Pi_\Omega\!\Bigl(
x_i^\ell
\!+\!\Delta t \bigl(f(x_i^\ell)+g(x_i^\ell)u_\theta(x_i^\ell,s_\ell)\bigr)
\!+\!\sigma\sqrt{\Delta t}\,\xi_i^\ell
\Bigr),
\label{eq-rollout}
\end{equation}
where $i=1,\dots,n$, $\ell=0,\dots,L-1$, the $\xi_i^\ell$ are independent $\mathcal N(0,I_d)$ vectors independent of the initial states, $I_d$ is the $d\times d$ identity matrix, and $\Pi_\Omega$ is Euclidean projection onto the closed box. This numerical boundary rule is distinct from the periodic interpretation of \eqref{eq-sde}.

For simulation, choose evaluation times $t_k=s_{\ell_k}$, where $1\le\ell_1<\cdots<\ell_{N_s}\le L$. The particle set $X_k=\{x_i^{\ell_k}\}_{i=1}^n$ is compared with target samples $Y_k$ and prescribed centers $C_k=\{c_{k,j}\}_{j=1}^{M_k}$. The entries of $Y_k$ are independent and identically distributed with density $\rho_k^{\mathrm{tar}}$, independently of the simulated particles. The formation-matching cost is
\begin{equation}
\mathcal L_{\mathrm{roll}}
=
\sum_{k=1}^{N_s} w_k
\left[
\MMD^2(X_k,Y_k)
\,+\,
\lambda_a \mathcal A_\tau(X_k,C_k)
\right],
\label{eq-l_roll}
\end{equation}
where $\MMD$ is the maximum mean discrepancy, the fixed weights satisfy $w_k\ge0$ and $\sum_k w_k>0$, and $\lambda_a>0$ balances center assignment against distribution matching. MMD compares unlabeled samples without estimating a density or assigning individual particle correspondences. We use the Gaussian kernel
\[
K_\gamma(x,y)=\exp\!\left(-\frac{\|x-y\|^2}{2\gamma^2}\right),
\qquad \gamma>0,
\]
where $\gamma$ is the prescribed spatial bandwidth, distinct from the diffusion coefficient $\sigma$. For sample sets $X=\{x_i\}_{i=1}^{n_X}$ and $Y=\{y_j\}_{j=1}^{n_Y}$, we define
\[
\begin{aligned}
\MMD^2(X,Y)
&=\frac{1}{n_X^2}\sum_{i,i'=1}^{n_X}K_\gamma(x_i,x_{i'})\\
&\quad+\frac{1}{n_Y^2}\sum_{j,j'=1}^{n_Y}K_\gamma(y_j,y_{j'})\\
&\quad-\frac{2}{n_Xn_Y}\sum_{i=1}^{n_X}\sum_{j=1}^{n_Y}K_\gamma(x_i,y_j),
\end{aligned}
\]
including diagonal pairs. This cost is differentiable with respect to particle positions and invariant to sample ordering. At the population level, Gaussian-kernel MMD is zero only for identical probability distributions \cite{Gretton2012}, making it suitable for matching multimodal formations.

For $X=\{x_i\}_{i=1}^n$ and $C=\{c_j\}_{j=1}^M$, the formation-center term combines a weighted distance cost with a penalty for uneven allocation:
\begin{align}
\pi_{ij}
&=
\frac{\exp(-\|x_i-c_j\|^2/\tau)}
{\sum_{r=1}^M \exp(-\|x_i-c_r\|^2/\tau)},
\\
\mathcal A_\tau(X,C)
&=
\frac{1}{n}\sum_{i=1}^n \sum_{j=1}^M \pi_{ij}\|x_i-c_j\|^2\notag\\
&\quad+\lambda_{\mathrm{occ}}\,\KL(\hat \pi \,\|\, \mathrm{Unif}),
\end{align}
where $\tau>0$ controls the concentration of the assignment weights $\pi_{ij}$, $\lambda_{\mathrm{occ}}>0$ is the allocation penalty weight, $\hat \pi_j=\frac{1}{n}\sum_i \pi_{ij}$ is the average assignment weight for center $j$, and $\mathrm{Unif}=(1/M,\dots,1/M)$. The relative entropy is $\KL(\hat\pi\|\mathrm{Unif})=\sum_{j=1}^M\hat\pi_j\log(M\hat\pi_j)$.

Using the $L^2(Q)$ residual penalty as a surrogate for the weak residual, the composite synthesis objective is
\begin{equation}
\begin{aligned}
\mathcal L
&=
\lambda_{\mathrm{dyn}}\|\mathcal F[\rho_\theta,u_\theta]\|_{L^2(Q)}^2
+\lambda_0\varepsilon_0^2
\\
&\quad
+\lambda_{\mathrm{flux}}\varepsilon_{\mathrm{flux}}^2
+\lambda_{\mathrm{roll}}\mathcal L_{\mathrm{roll}}
\\
&\quad
+\lambda_u \int_Q \|u_\theta\|^2 \rho_\theta\,dx\,dt .
\end{aligned}
\label{eq-full_loss}
\end{equation}
The weights $\lambda_{\mathrm{dyn}},\lambda_0,\lambda_{\mathrm{roll}},\lambda_u$ are positive, and $\lambda_{\mathrm{flux}}\ge0$ weights the bounded-domain penalty. Numerical optimization approximates the integrals in \eqref{eq-full_loss} by quadrature or sampling. The particle term specifies the formation objective, while the residual terms promote consistency of the density approximation with the controlled dynamics.

\subsection{Offline Optimization and Online Feedback Execution}
Formation mismatch is evaluated at selected times rather than every integration step. For static targets, we evaluate formation mismatch at $t=T/2$ and $t=T$; for mode splitting or aggregation, we add one intermediate time before the final approach. No full reference trajectory is prescribed for individual particles.

Each offline iteration samples residual-evaluation points and simulates particles to evaluate \eqref{eq-full_loss}. The PDE residual depends on the normalized density $\rho_\theta$ and the feedback $u_\theta$, so its gradients update both output maps and the common map. The particle formation cost depends on the feedback composition $U_\theta\circ T_\theta$: it directly updates $U_\theta$ and $T_\theta$, and changes the density approximation indirectly through $T_\theta$. Online, each agent applies $u_i(t)=U_\theta(T_\theta(x_i(t),t))$ with fixed $\theta$: one evaluation of the common map and control output map per agent and control update. Algorithm~\ref{alg-training} summarizes the procedure.

\begin{algorithm}[t]
\caption{Offline optimization and online feedback for SR-NFPC}
\label{alg-training}
\begin{algorithmic}[1]
\STATE Sample interior, initial, and boundary points for residual evaluation.
\STATE Evaluate $\rho_\theta$, $u_\theta$, and sampled PDE, initial, and boundary residual losses.
\STATE Simulate closed-loop particle trajectories with \eqref{eq-rollout} under $u_\theta$.
\STATE Compute formation-matching cost \eqref{eq-l_roll} from $\MMD$ and weighted center assignments.
\STATE Update $\theta$ using the sampled objective corresponding to \eqref{eq-full_loss}; repeat steps 1--5 for the prescribed iterations.
\STATE Fix $\theta$ and apply $u_i(t)=U_\theta(T_\theta(x_i(t),t))$ online.
\end{algorithmic}
\end{algorithm}

\section{Density Approximation Bound}
\label{sec-analysis}
The normalized network output $\rho_\theta$ approximates the exact density $\bar\rho_\theta:=\rho^{u_\theta}$ induced by the same feedback law. Both objects are defined on the periodic domain in this section. The analysis uses the Aubin--Lions compactness theorem \cite{SimonCompactness} and the Lions--Magenes time-continuity and energy identities \cite[Ch.~III, Sec.~1]{LionsMagenes1972}.

\begin{proposition}
\label{prop-wellposed}
Under the standing analytical assumptions, for every $u\in\mathcal U_{\rm ad}$, \eqref{eq-fpe} with initial density $\rho_0$ admits a unique weak solution satisfying
\[
\begin{aligned}
\rho^u&\in L^2(0,T;H^1(\Omega))\cap C([0,T];L^2(\Omega)),\\
\partial_t\rho^u&\in L^2(0,T;H^{-1}(\Omega)).
\end{aligned}
\]
Moreover, $\rho^u(\cdot,t)\ge0$ almost everywhere and $\int_\Omega\rho^u(x,t)\,dx=1$ for every $t\in[0,T]$.
\end{proposition}
\noindent\emph{Proof:} Omitted due to page limits.

Fix $u_\theta\in\mathcal U_{\rm ad}$ and a normalized periodic approximation $\rho_\theta$ satisfying
\begin{equation}
\rho_\theta\in L^2(0,T;H^1(\Omega)),\qquad
\partial_t\rho_\theta\in L^2(0,T;H^{-1}(\Omega)).
\label{eq-approx_regularity}
\end{equation}
This regularity gives a representative in $C([0,T];L^2(\Omega))$, so the time traces in the error estimate are well defined.
\begin{theorem}
\label{thm-residual}
Under the standing analytical assumptions and \eqref{eq-approx_regularity}, there is $C_\rho>0$, independent of $\theta$, such that
\begin{equation}
\begin{aligned}
&\sup_{t\in[0,T]}\|\rho_\theta(t)-\bar\rho_\theta(t)\|_{L^2(\Omega)}\\
&\quad+\|\rho_\theta-\bar\rho_\theta\|_{L^2(0,T;H^1(\Omega))}
\le C_\rho(\varepsilon_{\mathrm{dyn}}+\varepsilon_0).
\end{aligned}
\label{eq-state_bound}
\end{equation}
\end{theorem}
\noindent\emph{Proof:} Omitted due to page limits.

The constant $C_\rho$ depends only on $\Omega,T,\sigma,\|f\|_\infty,\|g\|_\infty$, and $M_u$. Theorem~\ref{thm-residual} concerns the periodic continuum model; sampling, time-discretization, and bounded-domain boundary errors are outside its scope.

\section{Experiments}
\subsection{Setup}
The experiments consist of a 1D Fokker--Planck (FP) transport benchmark and a 2D formation-control simulation. The 1D benchmark uses the domain $[-3,3]$, horizon $T=3$, time step $0.04$, diffusion $\sigma=0.12$, and $1800$ simulated particles. The target is a Gaussian profile with center $1.2$ and scale $0.09$, restricted to the domain and normalized to unit mass. We compare the proposed controller against a tuned mean-field planning baseline \cite{zhou2021large,YuLaiLiOsher2024MFP} and a tuned centralized sampled-goal baseline that assigns each agent a sampled terminal target and tracks it with centralized feedback.  

The formation control simulation uses the 2D domain $[-3,3]^2$, horizon $T=4$, time step $0.05$, diffusion $\sigma=0.15$, and $2500$ agents. We evaluate five target geometries, namely line, ring, split, transition, and triangle. We compare against a navigation-function potential-field baseline \cite{RimonKoditschek1992}, a consensus-style feedback baseline \cite{OlfatiSaber2006,RenBeard2008}, and the uncontrolled case. Metrics are terminal 2-Wasserstein distance $W_2$, control energy, and runtime.

\subsection{1D Planner Comparison}
\begin{figure}[t]
\centering
\includegraphics[width=0.95\columnwidth]{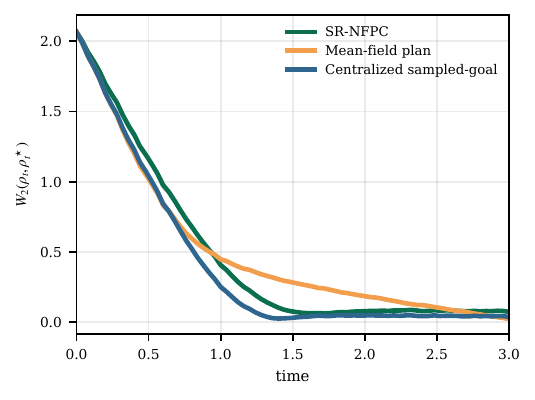}
\caption{Evolution of the 2-Wasserstein distance to the target distribution on the 1D benchmark. The plotted reference $\rho_t^\star$ is the fixed target density.}
\label{fig-planner_tracking}
\end{figure}

Fig.~\ref{fig-planner_tracking} compares the transient transport errors. After approximately $t=1$, SR-NFPC has a smaller discrepancy than the mean-field planner over most of the remaining horizon, whereas the planner attains the lowest terminal error. The curves complement the terminal metrics by showing how rapidly the population approaches the target distribution during the maneuver.

\begin{table}[t]
\centering
\caption{1D benchmark terminal metrics and total runtime.}
\label{tab-planner_compare}
\begin{tabular}{lccc}
\toprule
Method & $W_2$ & Energy & Time (s) \\
\midrule
SR-NFPC & 0.077 & \textbf{3.626} & 0.101 \\
Mean-field plan \cite{YuLaiLiOsher2024MFP} & \textbf{0.025} & 3.886 & 0.307 \\
Centralized sampled-goal & 0.040 & 5.077 & 0.044 \\
\bottomrule
\end{tabular}
\end{table}

Table~\ref{tab-planner_compare} shows that SR-NFPC trades terminal accuracy for lower control energy and shorter reported runtime relative to the mean-field planner. The centralized sampled-goal baseline provides an additional accuracy reference; its timing is not directly comparable because it uses a different communication architecture.

\subsection{Formation Benchmark}
\begin{figure}[t]
\centering
\includegraphics[width=0.8\columnwidth]{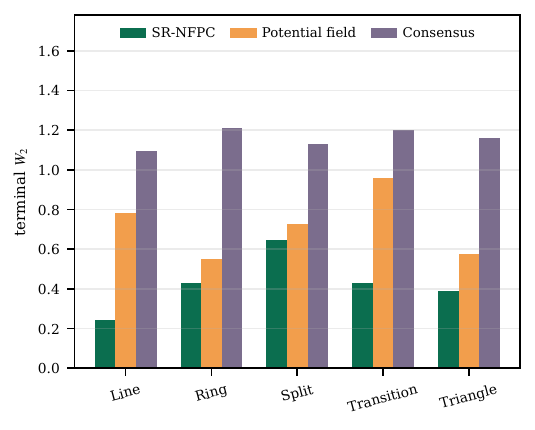}
\caption{Terminal Wasserstein distance across the five formation tasks. SR-NFPC achieves the lowest error among the compared methods on all five tasks.}
\label{fig-formation_w2}
\end{figure}

Fig.~\ref{fig-formation_w2} summarizes the five-formation suite. SR-NFPC achieves the lowest terminal Wasserstein distance among the compared methods on all five tasks, with particularly large reductions on the line, transition, and triangle targets.

\begin{table}[t]
\centering
\caption{Average performance over five formation tasks.}
\label{tab-formation_summary}
\begin{tabular}{lccc}
\toprule
Method & Avg.\ $W_2$ & Avg.\ energy & \shortstack{Best $W_2$\\count} \\
\midrule
SR-NFPC & \textbf{0.428} & 1.412 & \textbf{5/5} \\
Potential field \cite{RimonKoditschek1992} & 0.719 & 1.182 & 0/5 \\
Consensus \cite{OlfatiSaber2006,RenBeard2008} & 1.159 & 1.581 & 0/5 \\
No control & 0.931 & 0.000 & 0/5 \\
\bottomrule
\end{tabular}
\end{table}

Table~\ref{tab-formation_summary} makes the trade-off explicit. The proposed controller has the lowest average Wasserstein distance among the compared methods, with average control energy between those of the potential-field controller \cite{RimonKoditschek1992} and the consensus controller \cite{OlfatiSaber2006,RenBeard2008}. The no-control baseline confirms that diffusion alone is not enough to recover the target geometries.

\subsection{Effects of the Shared Representation}
Table~\ref{tab-arch_compare} shows that, on the transition task, replacing separate density/control networks with a shared representation reduces the parameter count from $17{,}154$ to $2{,}594$ while reducing terminal $W_2$ from $0.0542$ to $0.0394$ and runtime from $0.0862$\,s to $0.0621$\,s. This supports the design choice behind \eqref{eq-density_norm}--\eqref{eq-full_loss}, since the Fokker--Planck equation couples density and control directly.
 
\begin{table}[t]
\centering
\caption{Shared representation versus separate density/control networks on the transition task.}
\label{tab-arch_compare}
\begin{tabular}{lcc}
\toprule
Metric & Shared map & Separate maps \\
\midrule
Terminal $W_2$ & \textbf{0.0394} & 0.0542 \\
Control energy & \textbf{4.158} & 4.756 \\
Runtime (s) & \textbf{0.0621} & 0.0862 \\
Parameter count & \textbf{2594} & 17154 \\
\bottomrule
\end{tabular}
\end{table}

This comparison supports the use of shared coordinates: the shared model uses fewer parameters and achieves lower terminal $W_2$ and control energy on the reported task.

\section{Conclusions}
This paper presented a mean-field formation-transport method with direct online feedback evaluation based on a shared-representation neural Fokker--Planck controller. The combined objective uses density residuals to promote consistency with the controlled dynamics and particle costs to specify the target formation. The shared parameterization supports direct feedback evaluation after offline optimization. Future work will extend the controller to obstacle-rich environments, higher-dimensional formations, and stronger approximation guarantees for the particle-based formation cost.  

\bibliographystyle{IEEEtran}
\bibliography{refs}

\end{document}